\documentclass[11pt,a4paper]{article}
\usepackage[hyperref]{acl2021}
\usepackage{times}
\usepackage{latexsym}
\usepackage{enumitem}
\usepackage{paralist}

\usepackage{CJK}
\usepackage{amsmath}

\usepackage{microtype}

\aclfinalcopy 

\usepackage{graphicx} 
\usepackage{diagbox}
\usepackage{wrapfig}
\usepackage{xcolor}
\usepackage{amsmath}
\usepackage{amsfonts}

\usepackage{multirow}
\usepackage{booktabs}

\title{The Volctrans GLAT System:\\ Non-autoregressive Translation Meets WMT21}
\author{
  Lihua Qian\thanks{~~Equal contributions.}~~$^{\dag}$, Yi Zhou$^{*\dag}$, Zaixiang Zheng$^{*\dag}$, Yaoming Zhu$^{\dag}$, Zehui Lin$^{\dag}$, \\
  {\bf Jiangtao Feng$^{\dag}$, Shanbo Cheng$^{\dag}$, Lei Li$^{\ddag}$, Mingxuan Wang$^{\dag}$ and Hao Zhou$^{\dag}$ }\\
  $^{\dag}$Bytedance AI Lab \ $^{\ddag}$University of California Santa Barbara\\
  {\tt\small \{qianlihua,zhouyi.naive,zhengzaixiang,zhuyaoming,linzehui\}@bytedance.com}\\
  {\tt\small \{fengjiangtao,chengshanbo,wangmingxuan.89,zhouhao.nlp\}@bytedance.com}\\
  {\tt\small lilei@cs.ucsb.edu}
}

\date{}

\begin{document}
\begin{CJK}{UTF8}{gbsn}

\maketitle
\begin{abstract}
This paper describes the Volctrans' submission to the WMT21 news translation shared task for German$\rightarrow$English translation. 
We build a parallel ({\em i.e.}, non-autoregressive) translation system using the Glancing Transformer~\citep{glat}, which enables fast and accurate parallel decoding in contrast to the currently prevailing autoregressive models. 
To the best of our knowledge, this is the first parallel translation system that can be scaled to such a practical scenario like WMT competition. 
More importantly, our parallel translation system achieves the best BLEU score~(35.0) on German$\rightarrow$English translation task, outperforming all strong autoregressive counterparts.
\end{abstract}

\section{Introduction}

In recent years' WMT competitions, most teams develop their translation systems based on autoregressive models, such as Transformer~\citep{transformer}. 
Although autoregressive models (AT) achieve strong results, it is also worth exploring other alternative machine translation paradigm.
Therefore, we build our systems with non-autoregressive translation~(NAT) models~\cite{gu2017non}. 
Unlike the left-to-right decoding in the autoregressive models, the NAT models employ the more efficient parallel decoding. Specifically, our system employs single-pass parallel decoding, which generates all the tokens in parallel at one time, thus can accelerate decoding speed.

In this paper, we would like to present the best practice we explored in this year's competition for our parallel translation system, aiming at achieving top results while preserving decoding efficiency.

\paragraph{System Overview.} 
To achieve this, we improve the parallel translation system in several aspects, including better model architectures, various data exploitation methods, mutli-stage training strategy, and inference with effective reranking techniques.
For model architectures (\S\ref{sec:arch}), we build the parallel translation system based on the Glancing Transformer~\cite[GLAT,][]{glat}.
Besides, our system employs dynamic linear combination of layers~\cite[DLCL,][]{dlcl} for training deep models. 
For data exploitation (\S\ref{sec:data}), we first filter data with multiple strategies. After filtering, we use the Transformer~\citep{transformer} to synthesize various distilled data. 
For training (\S\ref{sec:training}), the NAT models employ multi-stage training to better exploit the distilled data. 
At inference phase (\S\ref{sec:inference}), the system generates the final results by reranking candidate hypothesis from multiple parallel generation models.

With the proposed techniques, our parallel translation system surpasses autoregressive models, and achieves the highest BLEU score~(35.0) in the German$\rightarrow$English translation task. Such results show that parallel translation system not only has great decoding efficiency, but also could achieve better performance compared to the autoregresssive counterparts. 

\section{Backbone Model Architecture}
\label{sec:arch}

\begin{figure}[t]
  \centering
  \includegraphics[width=.98\linewidth]{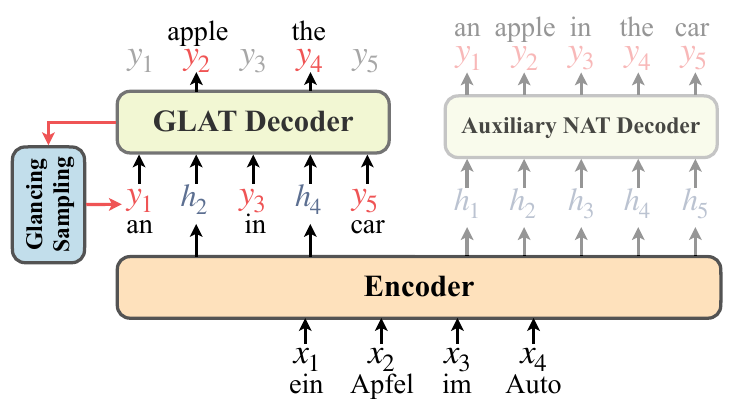}
  \caption{Illustration of our backbone model architecture: Glancing Transformer with an auxiliary decoder.
  }
  \label{fig:backbone}
\end{figure}

As depicted in Figure \ref{fig:backbone}, our submitted system employs GLAT~\citep{glat} as our backbone model architecture, and includes an auxiliary decoder in GLAT for achieving better translation performance.
GLAT is a method for training non-autoregressive models rather than a model architecture, which adaptively samples target tokens in training. 
Although the target token sampling in GLAT helps training, it also introduces a gap between training and inference.
To close the gap, we introduce the auxiliary decoder that shares the same encoder with the GLAT decoder, which is only used for training in a multi-tasking fashion.
Besides, we train models with three architecture settings to increase model diversity. 

\subsection{Glancing Transformer}
GLAT has three components: the encoder, the decoder, and the length predictor. The architecture of GLAT is built upon the Transformer~\citep{transformer}. The encoder is the same as that of Transformer, and the decoder is different from the Transformer decoder in the attention mask. Transformer employs attention mask in self-attention layer to prevent decoder representations attending to subsequent positions. Since GLAT generates sentences in parallel, the decoder of GLAT has no attention mask and uses global context in decoding. The details of the length predictor is described in Section~\ref{length}.

To reduce the difficulty of training deep models, we also employ dynamic linear combination of layers~\cite[DLCL,][]{dlcl} in the architecture. With DLCL, the input of each layer is the linear combination of outputs from all the previous layers.  

Given the source input $X=\{x_1,x_2,...,x_N\}$ and the target output $Y=\{y_1,y_2,...,y_T\}$, we use the glancing language model~\citep{glat} in training. The model performs two decoding during training. In the first decoding, the model generates the sentence $\hat{Y}$ in parallel. Then, the model randomly selects a subset of tokens $\mathbb{GS}(Y,\hat{Y})$ in the target sentence $Y$:
\begin{equation}
    \mathbb{GS}(Y,\hat{Y}) = \text{Random} (Y, S(Y, \hat{Y}))
\end{equation}
where $\text{Random}(Y, S)$ means randomly sample $S$ tokens in $Y$. And the sampling number $S(Y, \hat{Y})$ is computed by $S(Y, \hat{Y}) = \alpha \cdot d(Y,\hat{Y})$. $d(Y,\hat{Y})$ is the Hamming distance between the first decoding result $\hat{Y}$ and the target sentence $Y$, and $\alpha$ is a hyper-parameter for controlling the sampling number more flexibly.

In the second decoding, the model replaces part of the original decoder input representations with the embeddings of tokens in $\mathbb{GS}(Y,\hat{Y})$. Specifically, the token $y_i$ is used to replace the input representation at position $i$.
With the replaced decoder inputs, the model learns to predict the remaining words and compute the training loss:
\begin{equation}
    \mathcal{L}_{\text{glm}} = \sum_{y_t \in \overline{\mathbb{GS}(Y,\hat{Y})}}\\
     \log p(y_t| \mathbb{GS}(Y,\hat{Y}), X)
\end{equation}
where $\overline{\mathbb{GS}(Y,\hat{Y})}$ is the subset of tokens in $Y$ that are not selected.
In training, the model starts from learning to generate sentence fragments and gradually learning the parallel generation of the whole sequence.

\begin{figure*}[t]
  \centering
  \includegraphics[width = 12cm]{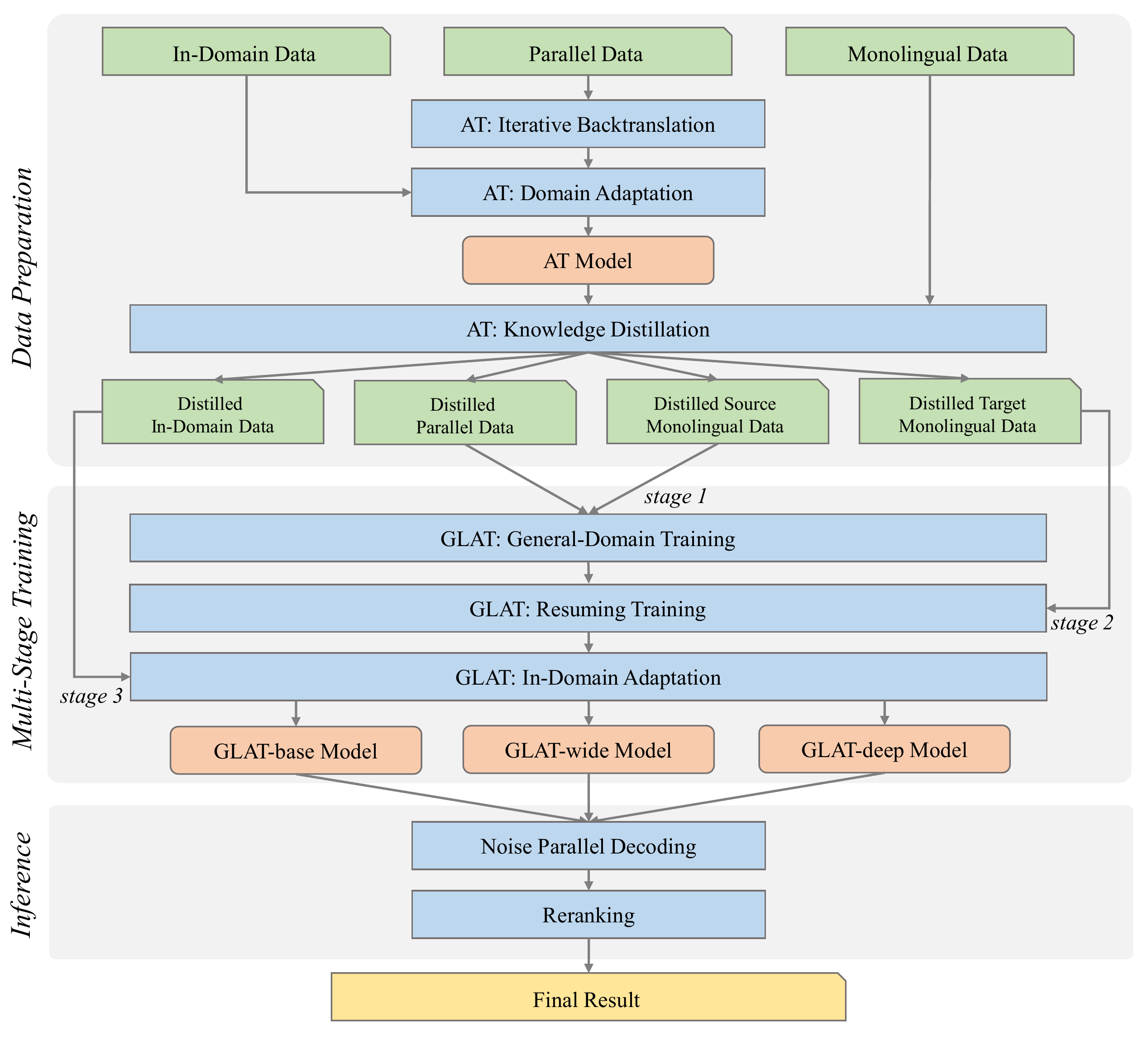}
  \caption{Overview of Volctrans GLAT System. Each grey block denotes a part of the system, the details can be found in Section \ref{sec:data}: Data Preparation, Section \ref{sec:training}: Multi-Stage Training, and Section \ref{sec:inference}: Inference. 
  }
  \label{fig:overview}
\end{figure*}

\subsection{Auxiliary Decoder}
Although the sampled target words in GLAT training help the model learn target word interdependencies, they also introduce a gap between training and inference as the model cannot obtain target word inputs in inference. Therefore, we add an auxiliary non-autoregressive decoder to close the gap. The auxiliary decoder shares the same encoder with the GLAT decoder and directly learns to predict the whole sequence in parallel. With the auxiliary decoder, we compute the loss for predicting the whole sequence:
\begin{equation}
    \mathcal{L}_{\text{aux}}=\sum_{t=1}^{T}\log P_{\text{aux}}(y_t|X)
\end{equation}
where $P_{aux}$ is the output probability of the auxiliary decoder. We jointly train the two decoders and the training loss of model is:
\begin{equation}
    \mathcal{L}_{\text{gen}}=\mathcal{L}_{\text{glm}}+\lambda \mathcal{L}_{\text{aux}}
\end{equation}
Note that the auxiliary decoder is only used in training and has no additional cost in inference. 
\subsection{Length Prediction}\label{length}
To enable parallel generation, the model predicts the target length before decoding. We use the average of encoder hidden states $H_{\text{avg}}$ as the representation to predict the length of target sentence. The probability of the target length is computed by:
\begin{equation}
    P_{\text{len}} = \mathrm{softmax}(H_{\text{avg}}^{\top} E_{\text{len}})\
\end{equation}
where $E_{\text{len}}$ is the embeddings of length. Instead of directly predicting the target length, the implemented model predicts the length difference between input and output, which is easier to learn. We use cross entropy loss for optimizing $P_{\text{len}}$ and train the length predictor with the generation module jointly.

\subsection{Model Variants}
\label{sec:mv}
As shown in Figure~\ref{fig:overview}, in order to increase the diversity of models, we use three model architecture settings for GLAT.
The details of the three GLAT architecture variants are:

\begin{compactitem}

    \item GLAT-base: Following~\citet{volctrans20};\citet{baidu19}, we increase the number of encoder layers and use 16 encoder layers for GLAT-base. For decoders, we use 6 layers for the original decoder and 2 layers for the auxiliary decoder. As for other model hyper-parameters, we use the 1024 hidden dimension and 16 attention heads, which are the same as the setting of Transformer-big. 
    \item GLAT-deep: We further increase the number of encoder layers to 32 for GLAT-deep. To keep the number of model parameters on the same scale, we decrease the hidden dimension to 768.
    \item GLAT-wide: Following previous work~\citep{volctrans20}, we also expand the dimension of the feed-forward inner layer to construct GLAT-wide. We set the feed-forward dimension to 12288 and the encoder layer number to 12.
    
\end{compactitem}

\section{Data Preparation}
\label{sec:data}

In this section, we will describe our best practice of distilled data construction by employing AT models. As illustrated in \textit{data preparation} in Figure \ref{fig:overview}, we will first depict the general procedure of data filtering and preprocessing of the provided raw data, followed by the training details of the AT models. Finally, we will describe how we produced distilled data given the trained AT models. The resulting distilled data will be used for training our GLAT system.

\subsection{Data Filtering and Preprocessing}

Data quality matters in machine translation systems. 
To obtain high-quality data, we employ rule-based heuristics, language detection, word alignment and similarity-based retrieval to filter the provided parallel and monolingual corpora.

\subsubsection*{Rule-based Data Filtering}
Based on experiences and WMT reports in previous years, we first preprocess raw data based on rules:
\begin{compactitem}
  \item Data deduplication. 
  \item Delete parallel data with the same source and target. 
  \item Remove special tokens and unprintable tokens.
  \item Remove HTML tags and inline URLs.
  \item Remove words or characters that repeat more than 5 times.
  \item Delete sentences that are too long (more than 200 words) or too short (less than 5 words), as well as the parallel data whose length-ratios of source and target sentences are out of balance.
\end{compactitem}

\subsubsection*{Parallel Data Filtering}
\label{sec:para}

After completing the rule-based filtering, we further filtered parallel data via language detection and its parallelism. 
The filtering process consists of three stages:
\begin{compactenum}
    \item Coarse-grained filtering: We filter parallel corpus according to the results and ratio of language detection. We use the pycld3\footnote{\url{https://pypi.org/project/pycld3/}} library to filter German$\rightarrow$English sentence pairs with a language likelihood greater than 0.8 and a language ratio greater than 60\%.
    \item Word alignment learning: We use \textit{fast align}~\cite{dyer2013simple}\footnote{\url{https://github.com/clab/fast\_align}} to automatically learn German$\rightarrow$English word alignment on the coarsely filtered corpus.
    \item Fine-grained filtering: We filter the sentences with an align score greater than five on all parallel corpora and sort them through the vocabulary learned by fast align.
\end{compactenum}
Note that the amount of data in different corpora is not balanced. We split the data into the paracrawl group and the non-paracrawl group. We filter out about 10\% of the data in the non-paracrawl group and 20\% of the data in the paracrawl group.

\subsubsection*{Monolingual Data Filtering}

For monolingual data, we first use the \textit{pycld3} library to filter the data of low scores, similar to the coarse-grained filtering of parallel data. 

Considering that monolingual data is too large, we searched for some of the most relevant sentences in our distilled data through sentence retrieval. 
We sample news domain sentences from the previous years' dev set and newscrawl corpus, and train a sentence BERT~\cite{reimers2019sentence}\footnote{\url{https://github.com/UKPLab/sentence-transformers}} to retrieve the sentences on the monolingual corpus. 
In detail, for each sampled news sentence, we calculate the inner product of sentence embedding between it and some random monolingual sentences (as the entire corpus is too large), where the sentence embedding is calculated with the sentence BERT model. 
We retrieved the top 8000 sentences for each news sample according to the inner product of sentence embedding. 
Finally, we deduplicate the retrieved sentences to obtain the final monolingual data. 

\subsubsection*{Data Preprocessing}
Once we obtained filtered data, we preprocess them through the following steps:
\begin{compactenum}
    \item Normalization: we use Moses tokenizer to normalize the punctuation.
    \item Tokenization: we use Moses tokenizer to tokenize all datasets. 
    \item Truecasing: we use Moses truecaser to learn and apply truecasing on all datasets.
    \item Subword segmentation: we use our proposed VOLT~\cite{xu2021volt}, which learns vocabularies via optimal transport, to split tokens into subwords, resulting in a joint vocabulary of a size of 12k subwords. 
\end{compactenum}
We summarize the statistics of the final datasets in Table~\ref{tab:data}.

\begin{table}[]
\centering
\begin{tabular}{@{}lcc@{}}
\toprule
                 & German (De) & English (En) \\ 
\midrule
parallel data    &            \multicolumn{2}{c}{75M}             \\
\midrule
monolingual data &      86M       &         105M     \\
\bottomrule
\end{tabular}
\caption{Statistics of the training data after preprocessing and filtering. }
\label{tab:data}
\end{table}

\subsection{Training of AT Systems}
In this section, we describe our AT systems, which served to distill data for GLAT training.
Overall, we first train a pair of German$\rightarrow$English and English$\rightarrow$German AT systems purely using parallel data. 
We then exploit source and target monolingual data to create synthetic parallel data to further improve the AT models.
Besides, we leverage the testsets from previous years to fine-tune the AT models for in-domain adaptation.

\paragraph{Hyperparameters.}
The AT models are Transformer models with 12 layers of encoder and decoder.
We use the implementations in Fairseq~\cite{ott2019fairseq}. All models are trained with Adam optimizer~\cite{kingma2014adam}. We use the inverse sqrt learning rate scheduler with 4000 warm-up steps and set the maximum learning rate to $5\cdot10^{-4}$. The betas are (0.9, 0.98).
We use multiple GPUs during training, resulting in an approximate total effective batch size of 128k tokens.
During training, we employ label smoothing~\cite{szegedy2016rethinking} of 0.1 and set dropout rate~\cite{srivastava2014dropout} to 0.3.

\subsubsection*{Iterative Back Translation}

\citet{zhang2018joint} proposed an iterative joint training method for better usage of monolingual data from the source language (i.e., German) and target language (i.e., English). 
In each iteration, the German$\rightarrow$English model generates forward synthetic data from the German monolingual data, and the English$\rightarrow$German model generates backward synthetic data from the English monolingual data. 
Then, the German$\rightarrow$English and English$\rightarrow$German models are trained with the new forward and backward synthetic data to improve both models' performance, in which the target-side data are assumed to be the authentic ones from the monolingual corpus.
In the next iteration, the German$\rightarrow$English and English$\rightarrow$German models can generate synthetic data with better quality, and their performance can be further improved . 
We jointly train the German$\rightarrow$English and English$\rightarrow$German models for $3$ iterations.

\subsubsection*{In-domain Finetuning}

We fine-tune the trained model on the previous years' testsets to obtain in-domain knowledge, which is a widely used technique in previous years' WMT~\cite{li-etal-2019-niutrans}. 
Specifically, we use WMT19 German$\rightarrow$English testset as in-domain data. 
We set the learning rate to 1e-4 without a learning rate scheduler and the max tokens per batch as 4096. 
We then fine-tune the model for 30 steps\footnote{Since the size of the in-domain data is small, fine-tuning with more steps will overfit the data.}.

\subsubsection*{Forward Translation}
\citet{bogoychev2019domain} observed that on the sentences that are originally in the source language, which is the case of the test sets of this year's WMT, the forward translation could bring significantly more improvement than back-translation.
We thus use the finetuned model, obtained by the aforementioned in-domain finetuning, to translate source monolingual corpus to obtain forward translation data. 
We then apply these forward translation data to finetune our AT models.

Finally, we combine all the parallel data, back-translation data, and forward translation data to further finetune our AT models. Table~\ref{tab:at} shows the performance of the AT models with respect with each training stage.
The resulting AT models are ready for constructing distilled data for GLAT training.

\begin{table}[]
\centering
\begin{tabular}{@{}lcc@{}}
\toprule
                 & De-En & En-De \\ 
\midrule
baseline    &          39.34       &        35.10           \\
iterative BT &          43.56     &         36.85                   \\
\midrule
in-domain FT &            44.00 &          38.30   \\
\midrule
forward translation &      44.05      &        39.50      \\
final training &          44.15       &        39.70 \\
\bottomrule
\end{tabular}
\caption{BLEU scores of AT models on \texttt{newstest20} with respect to different training stages.}
\label{tab:at}
\end{table}

\subsection{Constructing Distilled Data for GLAT}

One of the widely known difficulties of training NAT models is the multi-modality problem~\citep{gu2017non}. In the raw training data, the target tokens have strong correlations across different positions, which is hard to capture by NAT models due to the conditional independence assumption. 
A key ingredient in the training recipe for most of the NAT models is constructing training data via sequence-level knowledge distillation~\cite{kim2016sequence}, where the target-side of the training data is replaced by the forward translation of AT models. 

Note that previous work did not leverage existing large-scale monolingual data  in training GLAT models, either from source or target language.
In this work, we applied sequence-level knowledge distillation to parallel data and monolingual data from both source and target languages.
\begin{compactitem}
    \item Parallel data and source monolingual data distillation (119M sentences). We directly use German$\rightarrow$English AT model to obtain the forward translations of the German sentences.
    \item Monolingual target data distillation (39M sentences). The way to exploiting target monolingual data is not as evident as using the monolingual source data since the purpose of knowledge distillation is to construct a pseudo-parallel dataset where synthetic ones replace the actual target sentences. 
    To this end, we propose a cycle distilling technique. We use the backward English$\rightarrow$German AT model to back-translate the  monolingual target data, resulting in a translated source dataset. We then used the German$\rightarrow$English AT model to get the round-trip forward translation of the translated source dataset, obtaining the cycle distilled data. We will refer to this as \textit{cycle KD data}.
\end{compactitem}

\section{Multi-Stage Training}
\label{sec:training}
We train our parallel translation system in a multi-stage way (See \textit{Multi-Stage Training} in Figure \ref{fig:overview}). In the first stage, the model uses the distilled parallel and source monolingual data for training. In the second stage, we train the model with the target monolingual data (aka. \textit{cycle KD data}). After training the model on large-scale distilled data until convergence, we finetune the model on small-scale in-domain data. 


\begin{figure}[t]
  \centering
  \includegraphics[width = 7cm]{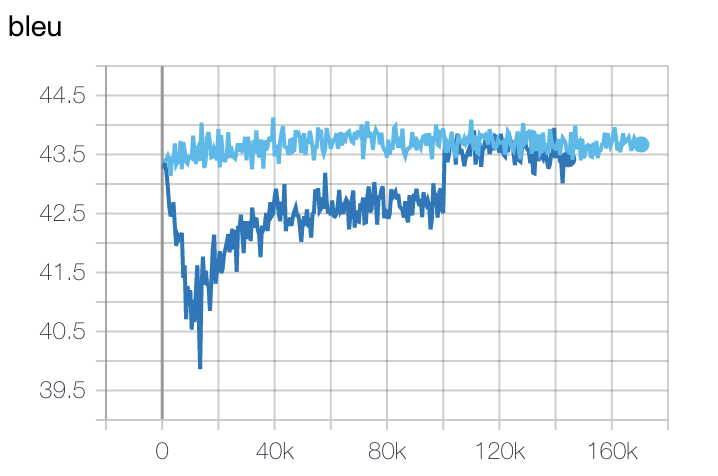}
  \caption{Learning curves of different finetuning strategies, reported on \texttt{newstest20}, De$\rightarrow$En. The light blue curve denotes training with inverse square root scheduler where the peak learning rate equals $5\cdot10^{-4}$,  and the initial sampling ratio $\lambda$ is set to $0.5$, the dark blue curve denotes training with a constant learning rate of $1e-4$ and $\lambda=0.1$.}
  \label{fig:nat_init_lr}
\end{figure}

\subsection{General-Domain Training}
All models are trained with Adam optimizer with decoupled weight decay~\cite{kingma2014adam,loshchilov2017decoupled}. We use the inverse sqrt learning rate scheduler with $4000$ warm-up steps and set the maximum learning rate to $5\cdot10^{-4}$. The adam betas are $(0.9, 0.999)$.

\begin{figure}[t]
    \centering
    \includegraphics[width = 6.5cm]{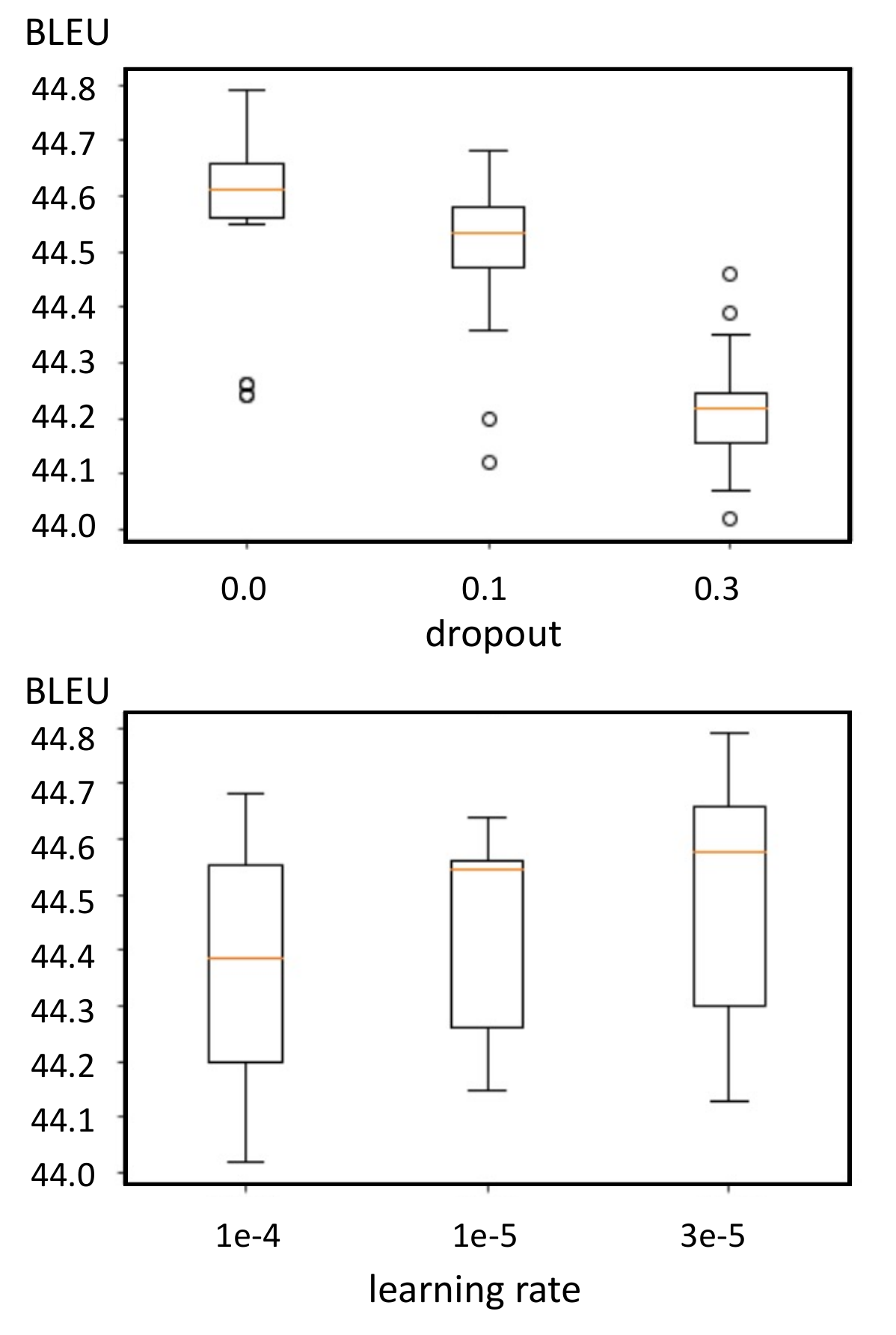}
    \caption{BLEU score versus the dropout and learning rate, reported on \texttt{newstest20}, De$\rightarrow$En.}
    \label{fig:nat_adapt_training}
\end{figure}

\subsection{Resuming Training}

We often have to load a pre-trained checkpoint and continuously train the model on a new dataset. The loaded checkpoint serves as a good initialization, and the parameters may change significantly in this process. 

We found that it is not easy to apply the techniques from auto-regressive translation to GLAT directly. Preliminary experiments show that if we employ the techniques illustrated in~\cite{glat} during the finetuning stage, the BLEU score will degrade dramatically and then increase slowly until convergence. The number of total update steps required for convergence is similar to training from scratch on a new dataset. There are mainly two concerns. Firstly, GLAT employs the inverse square root learning rate scheduler. The learning rate will increase to $5\cdot10^{-4}$ linearly and decay exponentially until the training process is over (the learning rate is close to $1e-4$). During the finetuning stage, a constant learning rate no larger than $1e-4$ will stabilize the training process. Secondly, the initial sampling ratio $\lambda=0.5$ in~\cite{glat} can be too large for finetuning since the model can already do a good job in the translation task. A large sampling ratio may cause the model to suffer from ``exposure bias''\cite{zhang2019bridging}: the gap between training (where some target words are provided) and validation (where no target words are provided). Figure \ref{fig:nat_init_lr} illustrates the comparison between two different finetuning strategies.

\begin{figure}[tbp]
  \centering
  \includegraphics[width = 7cm]{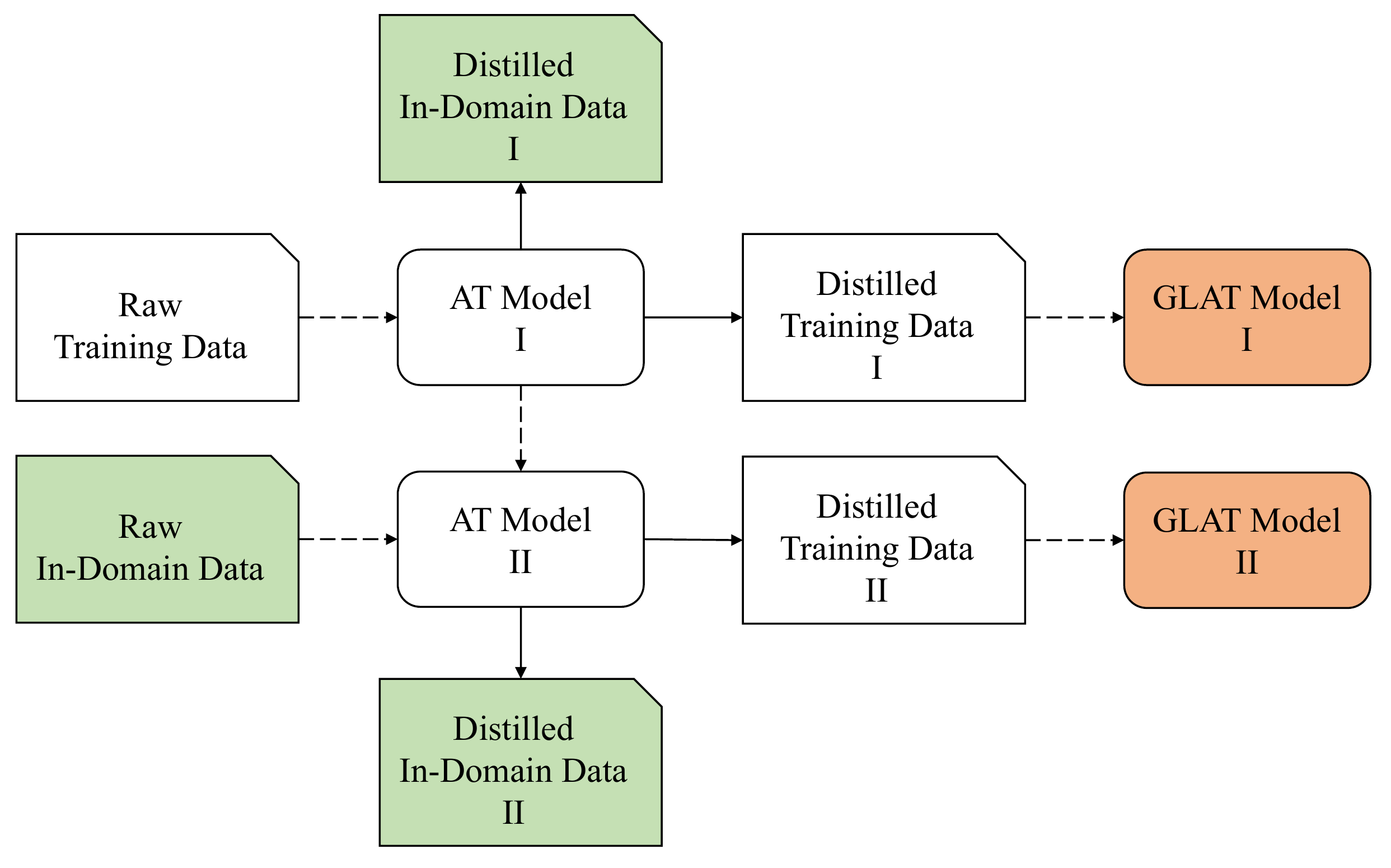}
  \caption{Various pipelines for domain adaptation.}
  \label{fig:nat_adapt_pipeline}
\end{figure}

\begin{table}[tbp] 
\begin{center}
\setlength{\tabcolsep}{1.0mm}
\begin{tabular}{ccc}
\toprule
{\bf In-Domain Data} & {\bf GLAT-I} & {\bf GLAT-II} \\ 
\midrule
- & $0.00$ & $+2.20$ \\
Raw & $+1.51$ & $+2.21$  \\
Distilled I & $+0.30$ & $+1.99$ \\
Distilled II & $+1.56$ & $+2.31$ \\
\bottomrule
\end{tabular}
\end{center}
\caption{ \label{tab:pipepline} Results of different adaptation pipelines. GLAT-I and GLAT-II are models trained with distilled training data generated by AT Model I and AT Model II in Figure~\ref{fig:nat_adapt_pipeline}, respectively. After training, we use the in-domain data Distilled I and Distilled II for fine-tuning.}
\end{table}

\subsection{In-Domain Adaptation}

When finetuning the model on small-scale in-domain data, which is widely used for domain adaptation~\cite{meng2020wechat}, the parameters of the model do not change significantly. 

For domain adaptation, we perform grid search on four group of hyper-parameters: learning rate( $1e-5$, $3e-5$, $1e-4$), dropout($0.0$, $0.1$, $0.3$), sampling rate $\lambda$ ($0.3$, $0.1$), and max number of tokens per batch ($2000$, $4000$, $8000$). For each combination, we conduct two experiments to reduce the variance. Experimental results (Figure \ref{fig:nat_adapt_training}) show that the learning rate and dropout rate are the most significant factors. Interestingly, when dropout is set to $0$, the performance is surprisingly great, which indicates the effectiveness of over-fitting on an in-domain dataset.

There are several feasible pipelines for domain adaptation due to the interaction between auto-regressive and non-autoregressive models. Figure \ref{fig:nat_adapt_pipeline} illustrates these pipelines, and the key points are listed as follows:

\begin{compactitem}
\item Should we finetune the auto-regressive model on the in-domain dataset (AT Model I$\rightarrow$AT Model II)? 
\item Should we use the original in-domain dataset for GLAT's model adaptation or the  in-domain dataset distilled by AT model I, or the in-domain dataset distilled by AT Model II?
\end{compactitem}

Table \ref{tab:pipepline} shows the results of different pipelines. Experiments show that making domain adaptation on the autoregressive model can boost the performance of the non-autoregressive model. It is also beneficial to further finetune the non-autoregressive model on the distilled in-domain dataset.

\section{Inference}
\label{sec:inference}

In this section, we introduce two approaches for GLAT's inference: Noisy parallel decoding (NPD) and Reranking (See \textit{Inference} in Figure \ref{fig:overview}). NPD is easy to integrate into a single model and improve the performance; Reranking can help push the performance to the limit: generating as many candidates as possible and ranking them with as many features as possible.

\begin{table}[tbp]
    \centering
    \begin{tabular}{cc}
    \toprule
        feature groups & feature number \\
    \midrule
        GLAT score & 3 \\
        AT 16e6d & 3 \\
        AT 12e12d & 3  \\
        Self BLEU & 1 \\
        Self Chrf & 1\\
    \bottomrule
    \end{tabular}
    \caption{Selected Features.}
    \label{tab:feat}
\end{table}

\begin{table} [tbp] 
\begin{center}
\setlength{\tabcolsep}{1.0mm}
\begin{tabular}{lccc}
\toprule
{\bf Model} & {\bf BLEU} & {\bf Self-R} & {\bf AT-R} \\ 
\midrule
 GLAT-base (w/o AUX)      & 42.28  & 42.54  & 42.90 \\
  + CTC & 41.04  & -      & -     \\ 
  + AUX  & 43.1   & 43.11  & 43.52 \\
\bottomrule
\end{tabular}
\end{center}
\caption{\label{tab:arch}Results of different architectures, reported on \texttt{newstest20}, De$\rightarrow$En. }
\end{table}

\begin{table*}[tbp] 
\begin{center}
\begin{tabular}{lccccccccc}
\toprule
& \multicolumn{3}{c}{\bf GLAT-base} 
& \multicolumn{3}{c}{\bf GLAT-deep} 
& \multicolumn{3}{c}{\bf GLAT-wide} \\
\midrule
{\bf Model} 
& {\bf BLEU} & {\bf Self-R} & {\bf AT-R} 
& {\bf BLEU} & {\bf Self-R} & {\bf AT-R}
& {\bf BLEU} & {\bf Self-R} & {\bf AT-R}\\ 
\midrule
  baseline
  & 43.10 & 43.11 & 43.52  & 42.44  & 43.89 & 43.14 & 43.38 & 43.49 & 43.81 \\
 + cycle KD 
  & 43.40 & 43.24 & 43.77  & 42.86  & 43.51 & 43.73 & 43.51 & 43.49 & 43.79 \\
 + adaptation 
  & 43.76 &	43.67 & 44.00  & 43.00  & 43.69 & 43.82 & 43.76 & 43.91 & 43.94\\
\midrule
 + reranker
& \multicolumn{9}{c}{44.64$^*$} \\

\bottomrule
\end{tabular}
\caption{\label{tab:nat_final}Final results, reported on \texttt{newstest20}, De$\rightarrow$En. $^*$ denotes the submitted system (BLEU=35.0 on \texttt{newstest21}, De$\rightarrow$En). The baseline is GLAT w/ AUX.}

\end{center}
\end{table*}

\subsection{Noisy Parallel Decoding}

A simple yet efficient inference approach is noisy parallel decoding (NPD)~\cite{gu2017non}. We first predict $m$ target length candidates (in Table \ref{tab:arch}, $m=5$), then generate output sequences with argmax decoding for each target length candidate. Then we use a model to rank these sequences and identify the best overall output as the final output. If the model for ranking and the one for generation is the same model (GLAT), we call it \textit{Self-Reranking}; if the ranking model is AT, we call it \textit{AT-Reranking}.

\subsection{Reranking}

We use kbmira\footnote{\url{https://github.com/moses-smt/mosesdecoder}} to re-rank hypotheses.
We first train GLAT model variants of different settings, each of which produces a set of candidates via the various search algorithm in Section~\ref{sec:mv}.
For each source sentence, every model outputs 7 hypothesis candidates and a total of 252 translations are collected for re-ranking.
Then we compute 44 features for each hypothesis, out of which 11 features are finally used. The selected features are listed in Table~\ref{tab:feat}.
The kbmira algorithm takes these features to select the best hypothesis from these candidates.
Note that the kbmira algorithm is optimized on \texttt{newstest19} and validated on \texttt{newstest20} to select the best feature combination.
Instead of enumerating all the possible combinations~($2^{44}$), we incrementally add feature groups to kbmira algorithm for fast search.

It is considered as an ablation study to pre-defined features.
After selecting the best feature combination, we further search better kbmira weights to achieve higher BLEU scores on \texttt{newstest20}.

\section{Experiment}
For our parallel translation system, we train three GLAT variants with the distilled data, and get the final outputs by reranking candidate hypothesis obtained from multiple GLAT models.

\subsection{Hyperparameters}
We implement our models with Fairseq~\citep{ott2019fairseq}. 
Our experiments are carried out on 4 machines with 8 NVIDIA V100 GPUs, each of which has 32 GB memory. The number of tokens per batch is set to $256k$. The dropout rate is set to $0.3$ for the first $100k$ steps. We reduce the dropout to $0.1$ after $100k$ steps, which can contribute to an improvement of about $1$ BLEU score (Figure \ref{fig:nat_init_lr}). The hyper-parameter $\lambda$ for balancing $L_{glm}$ and $L_{aux}$ is set to $1$.

\subsection{Results}
Our models are trained on the distilled parallel data and the distilled source monolingual data firstly. We experiment with various utilization of raw data, but the results show that the usage of raw data has no positive effect. The results of different architectures can be found in Table \ref{tab:arch}. Self-R and AT-R denote self-reranking and reranking with an autoregressive model, respectively. Experimental results show that the auxiliary decoder~(AUX) effectively improves the performance by about $0.6$ BLEU scores. For GLAT-base + CTC~\cite{graves2006connectionist}, we first set the max output length to twice the source input length and remove the blanks and repeated tokens after generation. We find CTC does not improve the performance and requires about twice the training time for convergence.

Based on GLAT with AUX, we employ three technologies to improve further: continuously training on the cycle KD data, domain adaptation, and reranking with various features. Table \ref{tab:nat_final} shows the final results of our submitted system. Training on the distilled target monolingual data can further improve the performance by about $0.3$ BLEU scores. Since the domain adaptation has already been employed in the AT model's training process, the cycle KD data has already contained information of the in-domain data. However, the domain adaptation on GLAT can still gain a slight improvement of about $0.2$. Moreover, an additional reranker with more diverse features can boost the performance by about $0.6$.

\section{Conclusion}
In this paper, we introduced our system submitted to the WMT2021 shared news translation task on German$\rightarrow$English. We build a parallel translation system based on the Glancing Transformer~\citep{glat}. Knowledge distillation, domain adaptation, reranking have proven effective in our system. Our constrained parallel translation system gets first place in the German$\rightarrow$English translation task with a 35.0 BLEU score.

\end{CJK}

\bibliographystyle{acl_natbib}
\bibliography{wmt21}

\end{document}